\documentclass[sigconf]{acmart}

\usepackage{booktabs} % For formal tables
\usepackage{multirow}
\usepackage{threeparttable}
\usepackage{flushend}

\setcopyright{acmcopyright}
\copyrightyear{2019}
\acmYear{2019}
\acmConference[XXXX 2019]{2019 XXXX}{XXX XX--XX, 2019}{XXXX, XXXX}
\acmPrice{15.00}
\acmDOI{XX.XXXX/XXXXXXX.XXXXXXX}
\acmISBN{XXX-X-XXXX-XXXX-X/XX/XX}

\begin{document}
\title{Ensemble Feature for Person Re-Identification}
%\titlenote{Produces the permission block, and
%  copyright information}
%\subtitle{Extended Abstract}
%\subtitlenote{The full version of the author's guide is available as
%  \texttt{acmart.pdf} document}

\author{Jiabao Wang, Yang Li, Zhuang Miao}
%\authornote{Jiabao Wang is the Corresponding author.}
\orcid{0000-0002-3706-9912}
\affiliation{%
  \institution{Army Engineering University of PLA}
  \streetaddress{Guanghua Road, Haifu Streat, No. 1}
  \city{Nanjing}
  \state{China}
  \postcode{210007}
}
\email{jiabao\_1108@163.com, solarleeon@outlook.com, emiao\_beyond@163.com}

\renewcommand{\shortauthors}{J. Wang et al.}

\begin{abstract}
In person re-identification (re-ID), the key task is feature representation, which is used to compute distance or similarity in prediction. Person re-ID achieves great improvement when deep learning methods are introduced to tackle this problem. The features extracted by convolutional neural networks (CNN) are more effective and discriminative than the hand-crafted features. However, deep feature extracted by a single CNN network is not robust enough in testing stage. To improve the ability of feature representation, we propose a new ensemble network (EnsembleNet) by dividing a single network into multiple end-to-end branches. The ensemble feature is obtained by concatenating each of the branch features to represent a person. EnsembleNet is designed based on ResNet-50 and its backbone shares most of the parameters for saving computation and memory cost. Experimental results show that our EnsembleNet achieves the state-of-the-art performance on the public Market1501, DukeMTMC-reID and CUHK03 person re-ID benchmarks.
\end{abstract}

\begin{CCSXML}
<ccs2012>
<concept>
<concept_id>10010147.10010178.10010224.10010240.10010241</concept_id>
<concept_desc>Computing methodologies~Image representations</concept_desc>
<concept_significance>500</concept_significance>
</concept>
</ccs2012>
\end{CCSXML}

\ccsdesc[500]{Computing methodologies~Image representations}

% We no longer use \terms command

\keywords{Person re-identification, ensemble feature, feature learning, convolutional neural networks.}

\maketitle

\section{Introduction}
\label{intro}
Person re-identification (re-ID) is an important task in computer vision and attracts lots of attention for its application in intelligent video surveillance. It aims to match pedestrians across different cameras. Due to the large variations in person appearance, pose, occlusion, illumination and so on, it is a very challenging problem. Fortunately, deep learning techniques have improved the performance effectively.
However, there is a big generalization gap~\cite{DBLP:journals/corr/KeskarMNST16} between training and testing. The main reason is that person re-ID problem is an unclose-set matching problem~\cite{DBLP:conf/cccv/WangLM17}, where the testing identities are different from the training ones. As we all known, classification problem in testing is to predict the label of the sample, which still belongs to the training labels. Different from classification, in person re-ID, the supervised labels of query person and gallery person both are not in the training set. As a result, it is difficult to learn effective features for person re-ID.

\begin{figure*}[!t]
\centering
\includegraphics[width=4.8in]{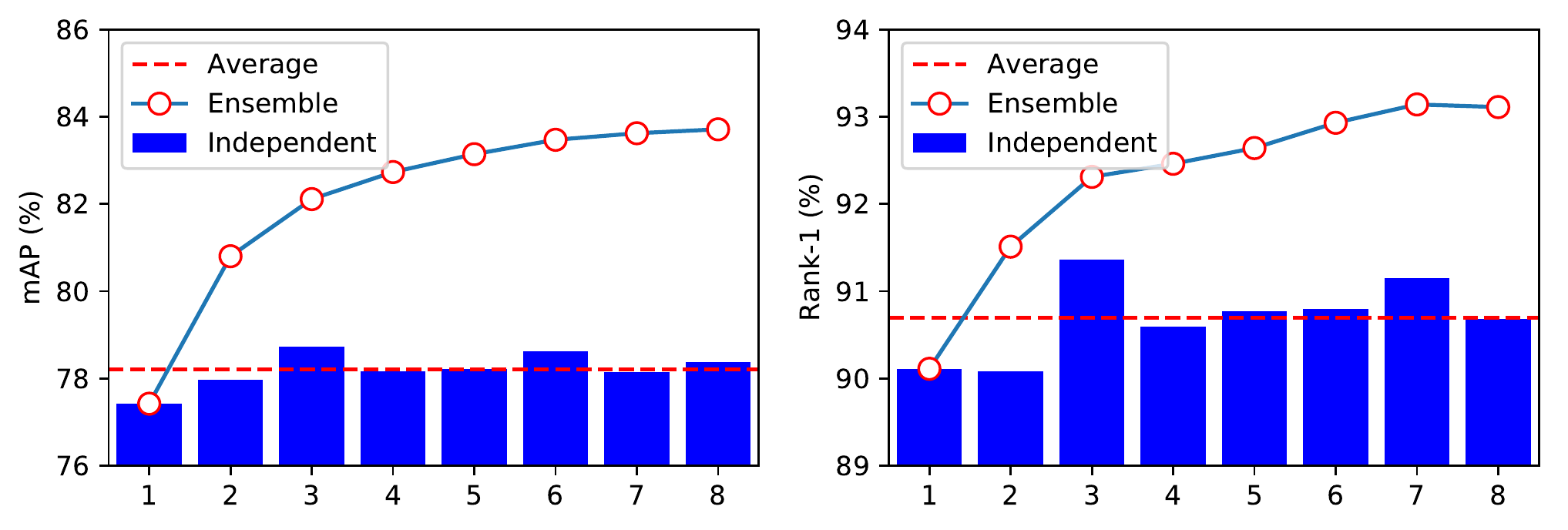}
\caption{Ensemble multiple networks. The networks are based on ResNet-50 backbone, we add a 1$\times$1 convolution to reduce the number of channels from 2048-dims to 256-dims, following softmax log-loss for classification. Note that the mAP and Rank-1 of independent networks are presented in histogram, while the mAP and Rank-1 of ensemble networks are showed in curve.}
\label{fig_independent1}
\end{figure*}

In a single classical CNN network, only one feature vector can be extracted for person re-ID, and it may have a limited feature representation ability. So it is possible to fuse multiple feature vectors to promote the representation ability and reduce the generalization gap. In practical, a simplest method is to just use multiple independent networks and concatenate their features to promote the performance. It's naturally an ensemble idea. To verify this simplest idea, we train 8 independent ResNet-50 networks on Market1501 dataset, and concatenate their features for person re-ID. The results are showed in Figure \ref{fig_independent1}, where we can find that the mAP and Rank-1 of 8 independent networks are similar and the average of them are showed in red dash line, while the ensemble features can easily achieve better results as the number of ensemble networks increases. However, it is inconvenient to manage multiple independent networks for deployment and the time-consuming increases linearly with the number of the networks.

In this paper, we explore to propose a new ensemble model with an end-to-end network with better generalization ability. The basic idea is create multiple branches to form multiple objectives. Each objective can be optimized to produce a solution for learning feature. Inspired by the part-based models~\cite{DBLP:journals/corr/abs-1711-09349}, we use different part-based model in each branch to make the features complementary. Finally, we evaluate the model and present a possible explanation.
The contributions of this work are as follows:
\begin{itemize}
  \item An ensemble network (EnsembleNet) is proposed to learn the feature representation for person re-ID. It's based on the ResNet-50 and consists of multiple branches. The features extracted from each branch are concatenated to form a feature for each image. It's an end-to-end architecture and has fewer parameters and computation than fusing multiple independent networks.
  \item To evaluate EnsembleNet, in experiments, we explore the effect of stride size, branch numbers and adaptive average pooling. The special setting can promote the performance effectively. Experimental results show that our approach achieves the state-of-the-art performance on the public Market1501, DukeMTMC-reID and CUHK03 person re-ID benchmarks.
  \item To explain the effectiveness of ensemble feature, inspiring by a visualization method of two-dimensional loss landscape~\cite{DBLP:journals/corr/abs-1712-09913}, we present the landscape of testing performance with the ``filter normalization''. The landscapes show that EnsembleNet has flatness of testing performance.
\end{itemize}

\section{Related Works}
Recently, person re-ID has attracted more attentions and achieved great improvements. Many of the existing works focus on feature learning and metric learning. Before deep learning becomes popular, there are many works explore to design hand-crafted features, such as local binary pattern (LBP)~\cite{DBLP:conf/eccv/XiongGCS14} features and local maximal occurrence (LOMO)~\cite{DBLP:conf/cvpr/LiaoHZL15} features. With the rise of deep learning, deep feature representation becomes the dominant methods and makes significant progress. The basic idea is to treat the person re-ID as a supervised distance metric learning problem. The traditional methods like keep it simple and straight forward metric (KISSME)~\cite{DBLP:conf/cvpr/KostingerHWRB12} and cross-view quadratic discriminant analysis (XQDA)~\cite{DBLP:conf/cvpr/LiaoHZL15} learn a transform matrix of features. Nowadays, in deep learning age, researchers pay more attention to design networks.

As we known, ResNet is the widely used model in person re-identification ~\cite{DBLP:journals/corr/ZhengYH16,DBLP:journals/corr/ZhengZY16,DBLP:conf/cccv/WangLM17}. In ResNet, the residual block is designed to tackle the gradient vanishing problem in the learning process. ResNet has a lot of paths from the input to the output with the short-cut connections and can be treated as a kind of ensemble model~\cite{DBLP:conf/cvpr/HeZRS16}. However, the network has only one loss and the parameters in the multiple paths are shared completely. The learned parameters just make the only one softmax log-loss to a minimum and are lack of diversity. Besides, GoogleNet is also a very successful model with multiple paths and losses to achieve great performance~\cite{DBLP:conf/cvpr/SzegedyLJSRAEVR15}. However, it just uses multiple paths to do convolution with kernels of multiple size, and the multiple loss with different length is designed for tackling the gradient vanishing problem. Compared with ResNet, the depth of GoogleNet limits its generalization ability. But we still can be motivated by the great ideas in design. To fit the special application of person re-ID, part-based models have become more important. Zhao et. al.~\cite{DBLP:conf/iccv/ZhaoLZW17} proposed the deeply-learned part-aligned representation for re-ID, while Sun et. al.~\cite{DBLP:journals/corr/abs-1711-09349} proposed a part-based convolutional baseline with a refined part pooling method. Both of them achieve the state-of-the-art performance and prove that part-based model is an effective method for person re-ID. These works also motivated us to design network with part-based idea.

For person re-identification, metric learning~\cite{DBLP:journals/corr/YiLL14,DBLP:conf/eccv/ShiYZLLZL16} is the main model for similarity ranking, which is introduced in the form of the contractive loss~\cite{DBLP:journals/corr/ZhengZY16} and triplet loss~\cite{DBLP:journals/corr/HermansBL17}, quadratic loss~\cite{DBLP:conf/cvpr/ChenCZH17}. Besides, the special designed losses, such as cosine loss~\cite{DBLP:conf/cccv/WangLM17} and sphere loss~\cite{DBLP:journals/corr/abs-1807-00537}, are also proposed to learn better feature representation. In practical, the combination of multiple losses can reach better performance. However, in this paper, we pay more attention to the design of the architecture, and we just use the simplest softmax log-loss for each branch in our network.

Inspired by these famous works, we design our EnsembleNet and evaluate its effects in following sections.

\section{EnsembleNet}
\subsection{Principle}

There are many works have been explored to ensemble deep networks ~\cite{DBLP:journals/taslp/ZhangW16a,DBLP:conf/ssci/ShaheenV16,DBLP:journals/spl/PandeyDMP17,Arsenovic18,DBLP:journals/access/HaS18} to improve the performance in different application domains. The typical methods just ensemble multiple independent deep networks to promote the results. However, there are two important weaknesses. One is that there are too many parameters in multiple networks and the numbers of the parameters increase linearly with the number of networks. The model is too large to deploy in practical. Another is that it is also time-consuming to train multiple deep models and the total time increases linearly with the number of networks. Even if the multiple GPU is used for parallel, there are also too cost. As a result, it has to decrease the complexity of the model when we design a deep ensemble model. Since multiple independent networks are too cost, it is possible to explore a new way to ensemble deep networks.

Recently, the part-based models become more popular for person re-ID~\cite{DBLP:journals/corr/abs-1711-09349,DBLP:journals/corr/abs-1803-10630}. Inspired by these models, we design a new end-to-end ensemble network, which has many branches to represent a person. In the designed network, each branch can be treated as a deep network, where most of the parameters have the same initialization with other branches. To design our model, we start from a traditional single network and break it into multiple branches, and each branch is an independent classifier for person re-ID. To promote the computing efficiency and representation ability, the designed model should obey the following three principles:
\begin{itemize}
\item The start point of the multiple branches should be selected elaborately, and the number of branches should be enough.
\item The feature should have semantic level representation ability, and its dimension should be low and can be merged easily.
\item The branches should be different from each other to have complement information.
\end{itemize}

Following the above principles, we design a deep ensemble network with the former part as a parameter-shared sub-network and the latter part as multiple parameter-independent branches. The former part share the parameters for saving memory and computation cost, and the latter part have multiple independent branches to reach the diversity. To save memory and computation cost, the start point of branches should be near to the rear of the network, while it should also be position front to keep the diversity of the representation. So we need to compromise memory and computation cost and the diversity of representation.

The ensemble network is designed by using a progressive strategy. There are three parts in the designed network. The first part is parameters-shared backbone sub-network. The second part is parameter-independent branches. The third part is a special designed module for classification. The whole network has several branches for ensemble representation.

The key problem is how to choose the start point of the branches. As the number of parameters and the robust of features are the two key points, so we set the start point of branches at the latter position of the sub-sampling layer. The reason is that most of the parameters are shared, and deep networks always have more abstract features when the size of the feature maps decreased.

Besides, another key problem is the number of the branches. To save the parameters, the number of the branches should be verified by experiments. The diversity of branches and the complexity of computation should be considered comprehensively. The detail of the network is depicted in the next sub-section.

\subsection{Architecture}

\begin{figure*}[!t]
\centering
\includegraphics[width=4.8in]{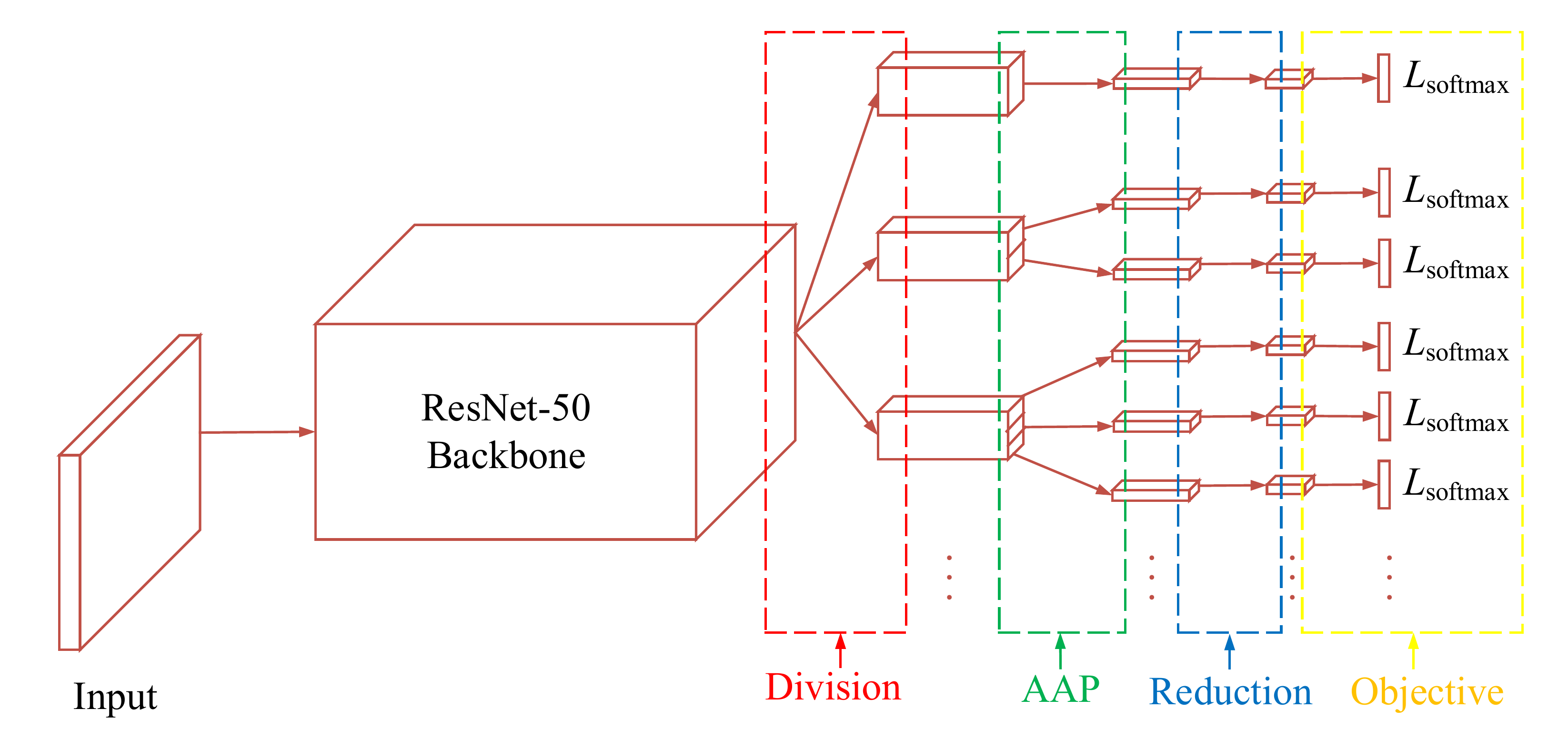}
\caption{Architecture of the proposed EnsembleNet.}
\label{fig_architecture}
\end{figure*}

The architecture is composed of multiple branches, based on the ResNet-50. Figure \ref{fig_architecture} shows the architecture of our proposed ensemble network (EnsembleNet). It can be divided into four parts: division module, adaptive average pooling (AAP) module, reduction module and objective module.

\paragraph{Division Module} Based on ResNet-50 backbone, the start point of branches is set at the \emph{res\_conv5\_1} layer, where there is a down-sampling operation. Before this layer, the backbone network has share most of the ResNet-50 parameters. At this layer, we break the backbone into $N$ branches, the parameters of which are independent from each other. After this layer, the advanced semantic information can be extracted from each branch.

\paragraph{AAP module} To represent the person as part model, we set different pooling operations for different branches. For the first branch, we use a global average pooling to get one feature vector of each image. For the second branch, we apply a 2D adaptive average pooling (AAP) over an input feature maps to get two vertical feature vectors for each image. For the third branch, we also adopt a 2D AAP over an input feature maps to get three vertical feature vectors for each image. We repeat the above operation for the next branches. For the $n$th branch, the output have $n$ feature vectors for each image. For $n$ branches, we can obtain $M=n \times (n+1)/2$ vectors for further computation.

Why the AAP is used for each branch? For the first branch, the global average pooling produces a global description of the person. For the second branch, the adaptive average pooling produces two vertical feature vectors for representing upper body and lower body. For the third branch, the adaptive average pooling produces three vertical feature vectors for representing head, body and legs. Different average pooling produces different number of features vectors for representing different local parts of a person. The features are extracted from different parts and have complemental characteristics.

\paragraph{Reduction Module} To represent the person, the features should be low-dimension for computing similarity. If we use the original 2048-dim features, the merged feature have $2048 \times M$ dimension. This is too cost for testing. So we reduce the dimension of features for each branch. The reduction module is composed of a $1 \times 1$ convolution to reduce the number of channels from 2048-dims to 256-dims, following batch normalization and leaky ReLU. The final feature has $256 \times M$, which is only the $1/8$ of the original feature dimension.

\paragraph{Objective Module} For each feature, we use a $1 \times 1 $ convolutional layer to replace the fully-connection layer, for mapping the feature to the number of classes. The softmax log-loss is used for classify the identities.
For each objective, the softmax log-loss can be computed from the feature $\mathbf{f}(I_i)$ of image $I_i$ and its truth label ${y}_{i}$. Each branch corresponds to one loss, which has the form of
\begin{equation}\label{eq1}
{{L}_{softmax}}=-\sum\limits_{i=1}^{B}{\log \frac{\exp (\mathbf{W}_{{{y}_{i}}}^{T}{\mathbf{f}(I_i)}+{{b}_{{{y}_{i}}}})}{\sum\nolimits_{j=1}^{C}{\exp (\mathbf{W}_{j}^{T}{\mathbf{f}(I_i)}+{{b}_{j}})}}}
\end{equation}
where $B$ is the mini-batch size, $C$ is the number of classes, and $\mathbf{W}_{j}$ and $b_{j}$ are the parameters to learn.

\subsection{Implementation}
To better learn the network, we initialize the parameters of the backbone and the branches by the parameters of the ImageNet pre-trained ResNet-50. Different branches are all initialized with the same pre-trained weights of the corresponding layers. The other parameters in the network are initialized by the `arxiver' method~\cite{DBLP:conf/iccv/HeZRS15}. The input image is uniformly resized to $384 \times 128$.

For training, the data augmentation is adopted. It includes random cropping, horizontal flipping and random erasing~\cite{DBLP:journals/corr/abs-1708-04896}. The mini-batch size of training is 32, and the examples are shuffled randomly. The SGD optimizer  is used with momentum 0.9. The weight decay factor is set to 0.0005. The learning rate is initialized from 0.01, and decay to 0.001 and 0.0001 after training for 40 and 60 epochs. The total training has 80 epochs. The learning rate of the parameters of the reduction module and the classifiers are 10 times learning rate of the pre-trained parameters.

For testing, we average the features extracted from an original image and its horizontal flipped one as the final feature. The cosine similarity is used for evaluating. Our model is implemented on Pytorch framework. It takes about 5 hours for training on Market1501 dataset with one NVIDIA GTX 1080TI GPU. To compare the performance of different methods, the two public evaluation metircs, CMC and mean Average Precision (mAP), are used. In all experiments, we use the single query mode and report the CMC at rank-1, rank-5, rank-10 and rank-20, and mAP~\cite{DBLP:conf/iccv/ZhengSTWWT15}.

\section{Experiments}
To evaluate the effectiveness of our ensemble model, we carry the experiments on three public datasets, including Market1501~\cite{DBLP:conf/iccv/ZhengSTWWT15}, DukeMTMC-reID~\cite{DBLP:conf/iccv/ZhengZY17}, CUHK03~\cite{DBLP:conf/cvpr/LiZXW14}. In all experiments, we abbreviate Market1501, DukeMTMC-reID and CUHK03 to Market, Duke, and CUHK. Market contains 701 identities of 12936 images for training, and 700 identities and clutter and background for testing. Duke is a subset of the DukeMTMC, and consists of 16522 images of 702 identities. CUHK contains 14096 images of 1467 identities which are captured from cameras in CUHK campus. The statistics of the datasets are presented in Table \ref{tab_datasets}. For Market and Duke, we use the standard evaluation protocol~\cite{DBLP:conf/iccv/ZhengSTWWT15}, while we use the new training and testing protocol for CUHK~\cite{DBLP:conf/cvpr/ZhongZCL17}.

% \usepackage{multirow}
\begin{table*}[]
\centering
\caption{Experimental datasets}
\label{tab_datasets}
\begin{threeparttable}
\begin{tabular}{|c|c|c|c|c|c|c|c|c|}
\hline
\multirow{2}{*}{\textbf{Datasets}} & \multicolumn{2}{c|}{\textbf{train}} & \multicolumn{2}{c|}{\textbf{gallery}} & \multicolumn{2}{c|}{\textbf{query}} & \multicolumn{2}{c|}{\textbf{total}} \\ \cline{2-9}
 & \textbf{images} & \textbf{ids} & \textbf{images} & \textbf{ids} & \textbf{images} & \textbf{ids} & \textbf{image} & \textbf{ids} \\ \hline
\textbf{Market} & 12936 & 751 & 19732 & 750* & 3368 & 750 & 36036 & 1501 \\ \hline
\textbf{Duke} & 16522 & 702 & 17661 & 1110 & 2228 & 702 & 36411 & 1404 \\ \hline
\textbf{CUHK} & 7365 & 767 & 5332 & 700 & 1400 & 700 & 14097 & 1467 \\ \hline
\end{tabular}
\begin{tablenotes}
    \footnotesize
    \item[1] `*' means that the number doesn't include the background and junks.
\end{tablenotes}
\end{threeparttable}
\end{table*}

\subsection{Comparison with state-of-the-arts}
To test the performance of EnsembleNet, we compare it with the state-of-the-art methods, such as IDE model~\cite{DBLP:journals/corr/ZhengYH16}, PAN~\cite{DBLP:journals/corr/ZhengZY17aa}, SVDNet~\cite{DBLP:conf/iccv/SunZDW17}, TriNet~\cite{DBLP:journals/corr/HermansBL17}, DaRe~\cite{DBLP:journals/corr/abs-1805-08805}, MLFN~\cite{DBLP:journals/corr/abs-1803-09132}, HA-CNN~\cite{DBLP:journals/corr/abs-1802-08122}, DuATM~\cite{DBLP:journals/corr/abs-1803-09937}, Deep-Person~\cite{DBLP:conf/icb/JinWLL17}, PCB~\cite{DBLP:journals/corr/abs-1711-09349}, Fusion~\cite{DBLP:journals/corr/abs-1803-10630}, SphereReID~\cite{DBLP:journals/corr/abs-1807-00537}. Results in details are presented in Table \ref{tab_state-of-the-arts}. we also show effects of re-ranking methods for improvement on mAP and Rank-1 accuracies. The results are divided into two groups, according to whether re-ranking is implemented or not.

\begin{table*}[!t]
\centering
\caption{Comparison with state-of-the-arts}
\label{tab_state-of-the-arts}
\begin{tabular}{|c|c|c|c|c|c|c|}
\hline
 & \multicolumn{2}{c|}{\textbf{Market}} & \multicolumn{2}{c|}{\textbf{Duke}} & \multicolumn{2}{c|}{\textbf{CUHK}} \\ \hline
\textbf{Methods} & \textbf{mAP} & \textbf{Rank1} & \textbf{mAP} & \textbf{Rank1} & \textbf{mAP} & \textbf{Rank1} \\ \hline
\textbf{IDE~\cite{DBLP:journals/corr/ZhengYH16}} & 50.7\% & 75.6\% & 45.0\% & 65.2\% & 19.7\% & 21.3\% \\
\textbf{PAN~\cite{DBLP:journals/corr/ZhengZY17aa}} & 63.4\% & 82.8\% & 51.5\% & 71.6\% & 34.0\% & 36.3\% \\
\textbf{SVDNet~\cite{DBLP:conf/iccv/SunZDW17}} & 62.1\% & 82.3\% & 56.8\% & 76.7\% & 37.3\% & 41.5\% \\
\textbf{TriNet~\cite{DBLP:journals/corr/HermansBL17}} & 69.1\% & 84.9\% & -- & -- & 50.7\% & 55.5\% \\
\textbf{DaRe(R)~\cite{DBLP:journals/corr/abs-1805-08805}} & 69.3\% & 86.4\% & 57.4\% & 75.2\% & 51.3\% & 55.1\% \\
\textbf{DaRe(De)~\cite{DBLP:journals/corr/abs-1805-08805}} & 69.9\% & 86.0\% & 56.3\% & 74.5\% & 50.1\% & 54.3\% \\
\textbf{MLFN~\cite{DBLP:journals/corr/abs-1803-09132}} & 74.3\% & 90.0\% & 62.8\% & 81.0\% & 47.8\% & 52.8\% \\
\textbf{HA-CNN~\cite{DBLP:journals/corr/abs-1802-08122}} & 75.5\% & 91.2\% & 63.8\% & 80.5\% & 38.6\% & 41.7\% \\
\textbf{DuATM~\cite{DBLP:journals/corr/abs-1803-09937}} & 76.6\% & 91.4\% & 64.6\% & 81.8\% & -- & -- \\
\textbf{Deep-Person~\cite{DBLP:conf/icb/JinWLL17}} & 79.6\% & 92.3\% & 64.8\% & 80.9\% & -- & -- \\
\textbf{PCB~\cite{DBLP:journals/corr/abs-1711-09349}} & 77.4\% & 92.3\% & 66.1\% & 81.7\% & 53.2\% & 59.7\% \\
\textbf{PCB+RPP~\cite{DBLP:journals/corr/abs-1711-09349}} & 81.6\% & 93.8\% & 69.2\% & 83.3\% & 57.5\% & 63.7\% \\
\textbf{Fusion~\cite{DBLP:journals/corr/abs-1803-10630}} & 79.1\% & 92.1\% & 64.8\% & 80.4\% & -- & -- \\
\textbf{SphereReID~\cite{DBLP:journals/corr/abs-1807-00537}} & 83.6\% & 94.4\% & 68.5\% & 83.9\% & -- & -- \\
\textbf{EnsembleNet} & 85.9\% & 94.8\% & 76.1\% & 87.1\% & 69.3\% & 73.8\% \\ \hline
\textbf{PAN+RR~\cite{DBLP:journals/corr/ZhengZY17aa}} & 76.6\% & 85.8\% & 66.7\% & 75.9\% & 43.8\% & 41.9\% \\
\textbf{TriNet+RR~\cite{DBLP:journals/corr/HermansBL17}} & 81.1\% & 86.7\% & - & - & 64.8\% & 64.4\% \\
\textbf{DaRe(R)+RR~\cite{DBLP:journals/corr/abs-1805-08805}} & 82.0\% & 88.3\% & 74.5\% & 80.4\% & 63.6\% & 62.8\% \\
\textbf{DaRe(De)+RR~\cite{DBLP:journals/corr/abs-1805-08805}} & 82.2\% & 88.6\% & 73.3\% & 79.7\% & 61.6\% & 60.2\% \\
\textbf{EnsembleNet+RR} & 93.0\% & 95.6\% & 88.0\% & 90.1\% & 81.7\% & 81.2\% \\ \hline
\end{tabular}
\end{table*}

\textbf{Results on Market}: From Table \ref{tab_state-of-the-arts}, without re-ranking, EnsembleNet achieved 94.8\% in Rank-1 and 85.9\% in mAP. After implementing re-ranking, the results are improved to Rank-1/mAP=95.6\%/ 93.0\%, which surpasses all existing methods. Among these methods, we note that the IDE model~\cite{DBLP:journals/corr/ZhengYH16} is a widely used baseline for deep re-ID systems. Our results exceed IDE by +19.2\% in Rank-1 and +35.2\% in mAP. SphereReID~\cite{DBLP:journals/corr/abs-1807-00537} and PCB+RPP~\cite{DBLP:journals/corr/abs-1711-09349}, which are two of the newest methods, achieves the second and third mAP without re-ranking. Our method exceeds them by +2.3\% and 4.3\% in mAP.

\begin{figure*}[!t]
\centering
\includegraphics[width=4.5in]{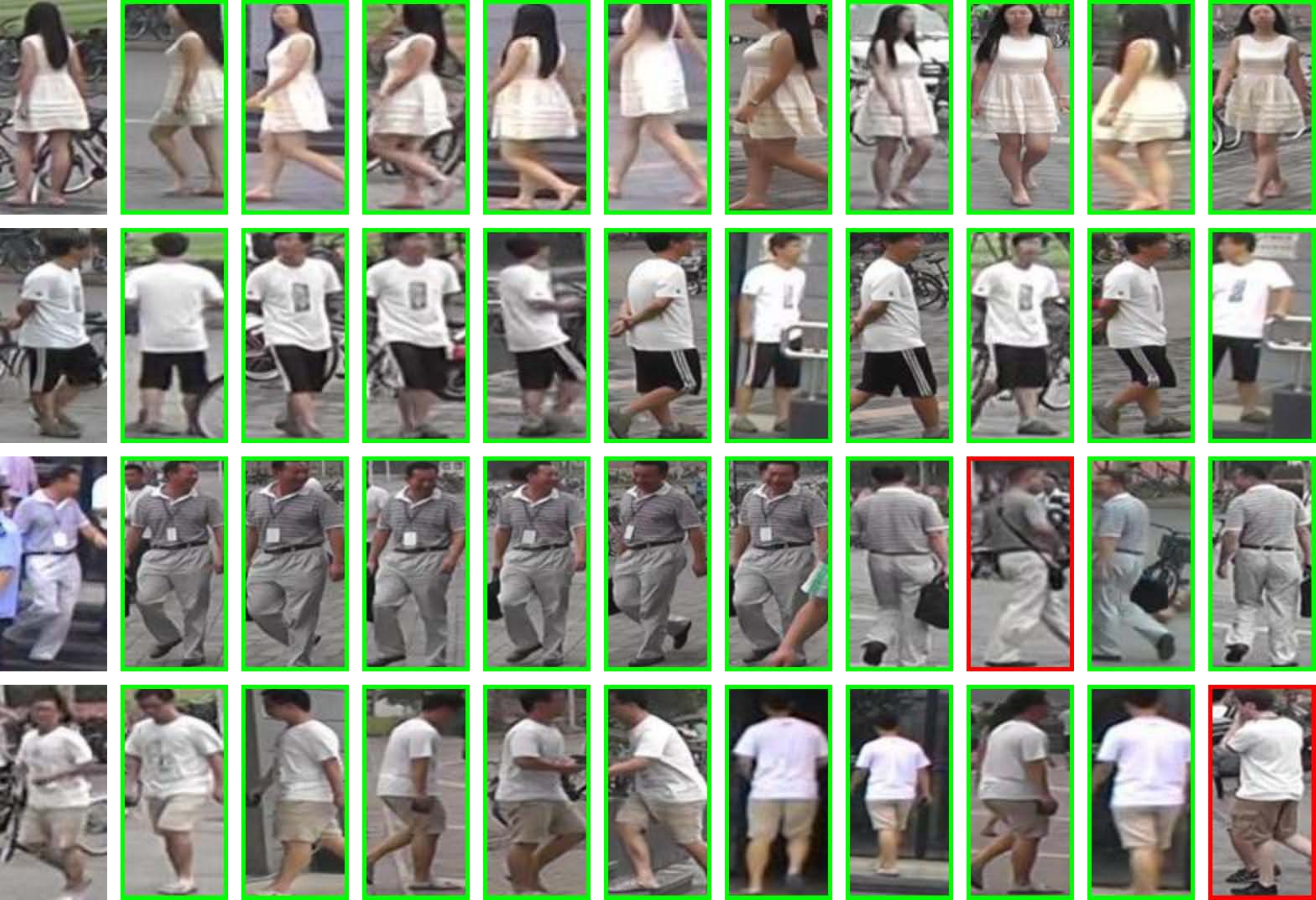}
\caption{Query examples.}
\label{fig_example}
\end{figure*}

Figure \ref{fig_example} shows top 10 ranking results for some given query pedestrian images. The first two results show the great robustness. Regardless of the pose or gait of these captured pedestrian, the ensemble features can robustly represent discriminative information of their identities. The third query image shows a man with an identity card in front of his chest, but we can retrieve his captured images in back view in rank-7, 9, 10. We attribute this surprising result to the ensemble feature, which have better generalization ability. The last query is captured in a low-resolution condition, losing an amount of important information. However, most of the ranking results are accurate and with high quality.

\textbf{Results on Duke}: The results achieved by our EnsembleNet show excellent performance. Without re-ranking, EnsembleNet achieves state-of-the-art result of Rank-1/mAP=87.1\%/76.1\%, outperforming the SphereReID with +3.2\%/+7.6\% in Rank-1/mAP. With re-ranking, EnsembleNet exceeds the second one with a large margin, and also achieves the best performance.

\textbf{Results on CUHK}: As presented in Table 2, without re-ranking, our EnsembleNet achieves Rank-1/mAP= 73.8\%/69.3\% on the detected setting, which outperforms the PCB+RPP by +10.1\% in Rank-1 and +11.8\% in mAP. After implementing re-ranking, our method outperforms all the published results by a large margin. It has to be noticed that the detected setting is harder than the labeled setting, because the detection failure has a great influence on re-ID.

\subsection{Abality Study}

\subsubsection{Stride Size}
In experiments, we find that the down-sampling operation (convolution with stride 2 at \emph{res5a}) has an obvious affection on the performance. If we replace the stride 2 as stride 1, we can get a higher performance. On Market, the mAP and Rank-1 with stride 1 are 80.23\% and 91.67\%, which has an improvement of +2.35\% and +0.84\%. Similar results are found on Duke and CUHK. The reason is that more information has been captured by convolution with stride 1 than that of convolution with stride 2. The output feature maps at \emph{res5a} have doubled size due to the stride 1, which does not reduce the resolution. As a result, the computation of the following layers is doubled. However, the number of parameters keeps the same as before. For application, we can replace stride 2 with stride 1 when we need higher performance. Furthermore, the stride 1 should be used when we have less computation resource. Note that the network with stride 1 is our \textbf{baseline} for further comparison.


\begin{table*}[!t]
\centering
\caption{Results on different strides}
\label{tab_stride}
\begin{tabular}{|c|c|c|c|c|c|c|}
\hline
\textbf{Datasets} & \textbf{\#Stride} & \textbf{mAP} & \textbf{Rank1} & \textbf{Rank5} & \textbf{Rank10} & \textbf{Rank20} \\ \hline
\multirow{2}{*}{\textbf{Market}} & {2} & 77.88\% & 90.83\% & 96.67\% & 97.86\% & 98.60\% \\ \cline{2-7}
 & {1} & 80.23\% & 91.67\% & 96.99\% & 98.24\% & 98.88\% \\ \hline
\multirow{2}{*}{\textbf{Duke}} & {2} & 66.25\% & 81.24\% & 90.44\% & 92.68\% & 94.97\% \\ \cline{2-7}
 & {1} & 68.60\% & 83.12\% & 91.38\% & 93.49\% & 95.42\% \\ \hline
\multirow{2}{*}{\textbf{CUHK}} & {2} & 56.96\% & 61.64\% & 78.86\% & 85.57\% & 90.64\% \\ \cline{2-7}
 & {1} & 58.67\% & 63.79\% & 79.86\% & 87.07\% & 91.86\% \\ \hline
\end{tabular}
\end{table*}

\subsubsection{Independent Ensemble}

\begin{figure*}[!t]
\centering
\includegraphics[width=3.6in]{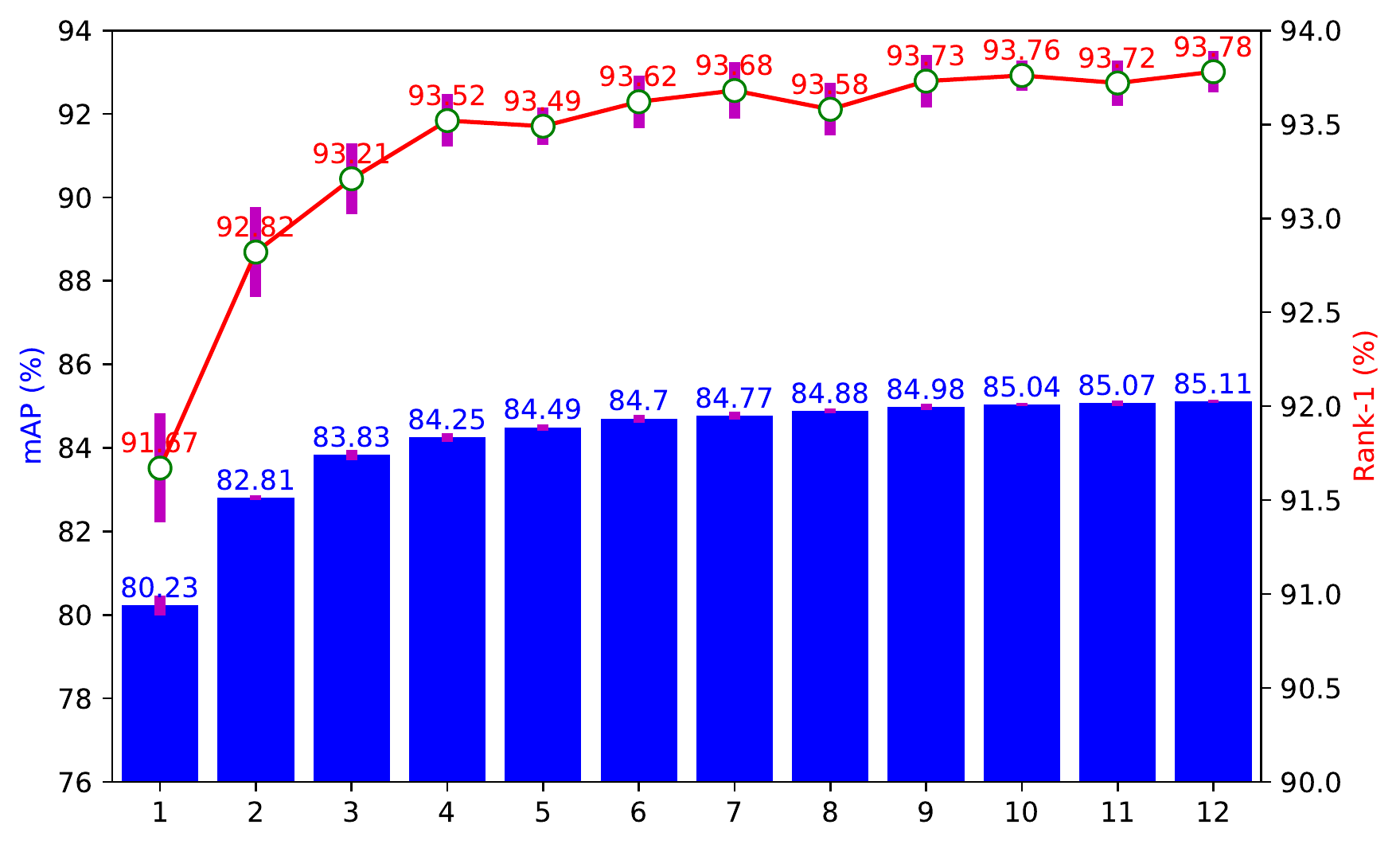}
\caption{Results on different number of independent networks.}
\label{fig_independent_ensemble}
\end{figure*}

To compare with the traditional ensemble methods, we train multiple networks independently and concatenate the features extracted by these networks. In experiments, 16 baseline networks are trained and the features are extracted from these networks. The performance is tested with different number of networks, the results are presented in Figure \ref{fig_independent_ensemble}. Given a specified number of features, we randomly selected the features and concatenate them to form an ensemble representation, and evaluate the mAP and Rank-1 performance. For each specified number, the evaluation process is repeated 10 times. In Figure \ref{fig_independent_ensemble}, the variance of each repeated process is expressed by a vertical red line segment. The long line segment represents the large variance. From the figure, we can find that the performance improves when the number of networks increases. The increasing rate slows down as the number increases, and tends to zero when the number of networks reaches 6. Furthermore, all the variance is very small, no more than 0.3. And the variance decreases as the number increases.

\subsubsection{Branch Numbers}
\begin{figure*}[!t]
\centering
\includegraphics[width=4.8in]{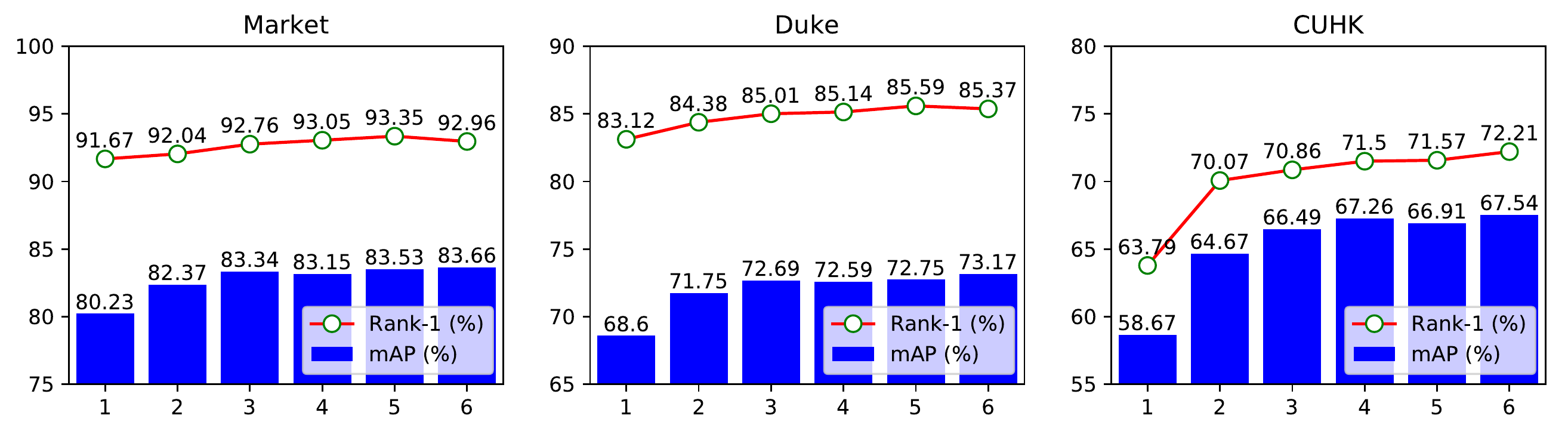}
\caption{Results on different branch numbers.}
\label{fig_branches}
\end{figure*}

To verify the performance in detail, we experiment the model with different number of branches. We test the branches from 1 to 6, where there is only one objective in each branch when AAP module is replaced with GAP module. The structures are similar while the difference is just the number of objectives. The performances are showed in Figure \ref{fig_branches}.

From Figure \ref{fig_branches}, it can be found that when the number of branches from 1 to 3, the improvement has changed greatly, and then the increasing speed has decreased and reaches a limit. When the number of branches achieves 6, the mAP/Rank1 of Market, Duke and CUHK are 83.66\%/92.96\%, 73.17\%/85.37\%, and 67.54\%/72.21\%. They have got an obvious improvement of +3.43\%/+1.29\%, +4.57\%/+2.25\%, and +8.87\%/+8.42\% from baseline. Considering both computation and performance, we can choose 3 branches for further study.

Besides, the method fusing multiple independent networks reaches better performance than the method with the same branches. For example, on Market dataset, the mAP of the former achieves 83.83\% when fusing three independent networks, while the latter achieves 83.34\%. The reason may be that the features extracted from the branches with the same backbone network may be redundancy.

\subsubsection{Adaptive Average Pooling}
\begin{figure*}[!t]
\centering
\includegraphics[width=4.8in]{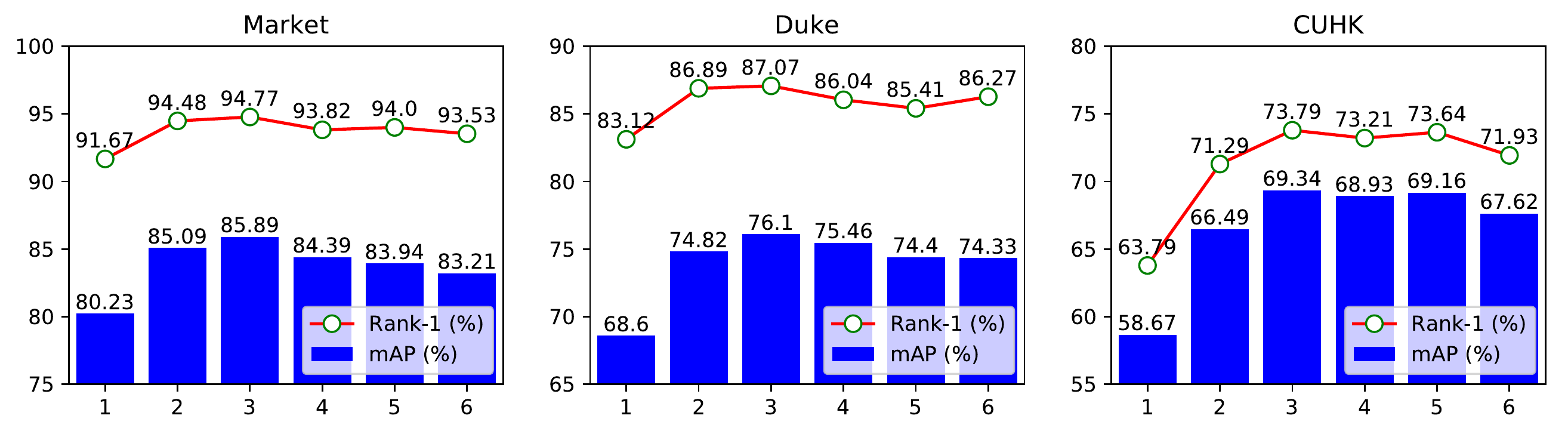}
\caption{Results on different objectives by AAP.}
\label{fig_app}
\end{figure*}

Furthermore, we divide the output of \emph{res5c} to different parts for each branch. We use adaptive average pooling (AAP) to make the feature maps into $N$ patches, and the division is done on the vertical. For different branches, we use different number of patches. For simplification, we add the patches as the number of branches increases. For $i$th branch, there are $i$ patches. Each patch corresponds to one objective, so there are $i \times (i+1)/2$ objectives. We test the performance of the different branches and patches. The results are shown in Figure \ref{fig_app}. From the figure, we can find that the mAP and Rank-1 achieves the best performance when the number of branches equals 3 (6 objectives). The corresponding mAP/Rank-1=85.89\%/94.77\% on Market, 76.1\%/87.07\% on Duke and 69.34\%/73.79\% on CUHK, which have an improvement of +5.66\%/+3.10\%, +7.50\%/+3.95\%, and +10.67\%/+10.00\% from baseline.

Comparing with independent ensemble method (Figure \ref{fig_independent_ensemble}), our EnsembleNet with AAP achieves better performance. As we known, the features extracted from independent ensemble reflects the appearance of the whole person. However, the AAP divides a person into different vertical parts and produces multiple local features. Each feature reflects an appearance of a local region, and the features are complemented with each other. So the mAP/Rank-1=85.89\%/94.77\% on Market, which has an improvement of 2.06\%/1.56\% from the method fusing three independent networks.

\section{Explanation of Effectiveness}
EnsembleNet is an ensemble model, which concatenates multiple features to achieves better representation based on an end-to-end multiple-branch and multiple-objective network. Why the concatenated multiple features can promote the performance?

To give a reasonable explanation, we resort to one standing hypothesis, flatness of minima. According to the work of~\cite{DBLP:journals/corr/KeskarMNST16}, the generalization gap between training and testing is related to the flatness of minimum of the objective. A flat minimum can achieve good generalization, while a sharp minimum can worse the generalization. As a result, if there is multiple minima, the flatness of minima of the objectives can be expanded. In~\cite{DBLP:journals/corr/KeskarMNST16}, flatness is characterized by the magnitude of the eigenvalues of Hessian, and $\epsilon$-flatness is defined as an approximation. However, Dinh et. al.~\cite{DBLP:conf/icml/DinhPBB17} shows that the measure of sharpness in~\cite{DBLP:journals/corr/KeskarMNST16} is problematic and redefinition is required for explaining generalization gap. The reason is that the sharpness can be easily manipulated through re-parametrization, when deep neural network is constructed with non-linearity rectifier. Fortunately, Li et. al.~\cite{DBLP:journals/corr/abs-1712-09913} propose a visualization method of two-dimensional loss landscape, based on ``filter normalization''. It can be used for explanation of the generalization. Based on this work, we present the landscape of different re-ID models for explanation.

\begin{figure*}[!t]
\centering
\includegraphics[width=4.8in]{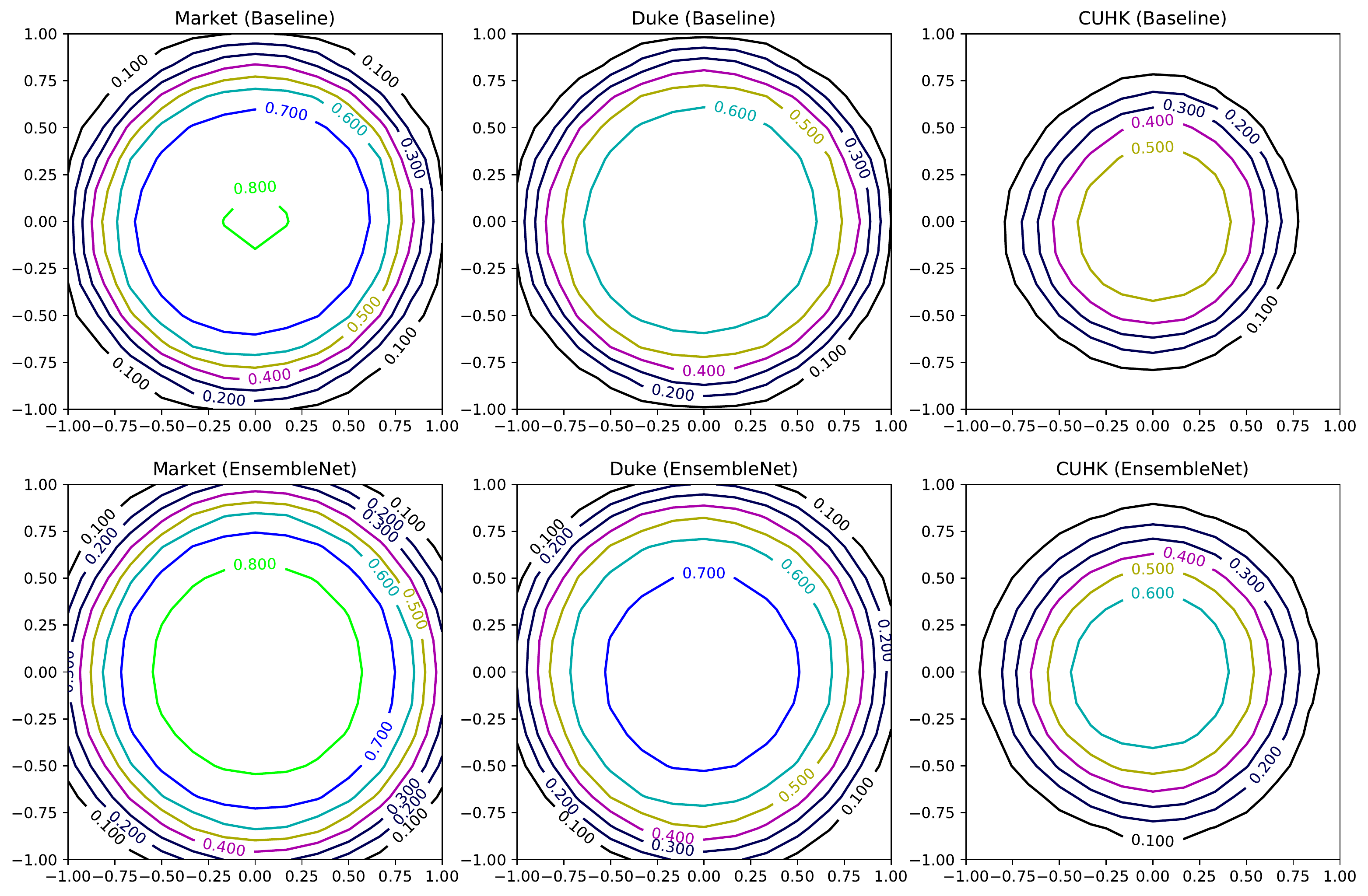}
\caption{The mAP landscapes of Baseline model and Ensembled Model.}
\label{fig_landscapes_map}
\end{figure*}

\begin{figure*}[!t]
\centering
\includegraphics[width=4.8in]{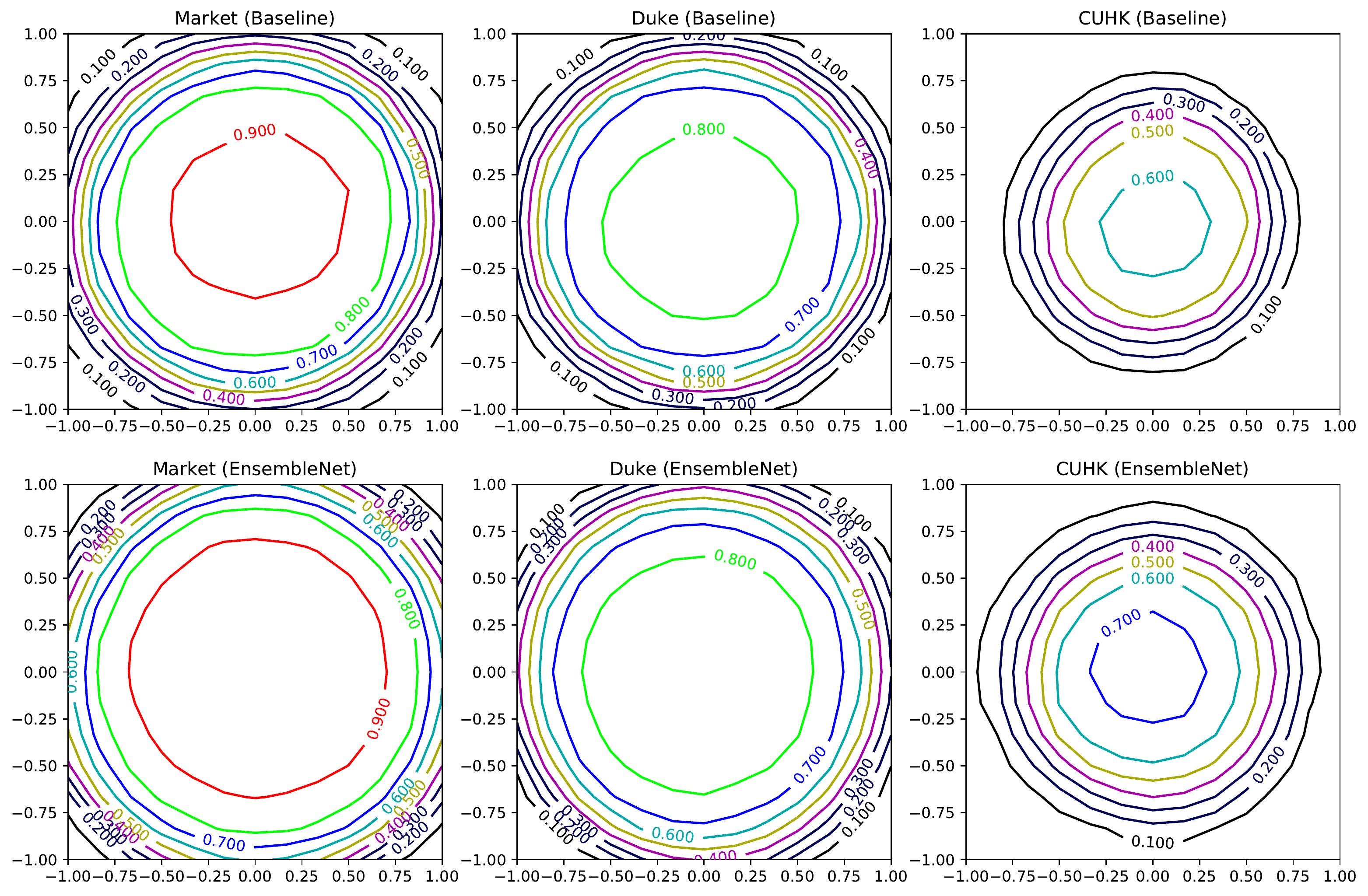}
\caption{The Rank-1 landscapes of Baseline model and Ensembled Model.}
\label{fig_landscapes_rank}
\end{figure*}

For classic CNN network, the task is always a non-convex optimization problem, so the solution is only one of the multiple minima. However, EnsemleNet has multiple objectives and the losses, it can reach multiple minima and is difficult to compare the landscapes of losses. So we turn our attention to the generalization ability, and present mAP and Rank-1 landscapes in Figure \ref{fig_landscapes_map} and Figure \ref{fig_landscapes_rank}. In Figure \ref{fig_landscapes_map} and Figure \ref{fig_landscapes_rank}, we can find that EnsembleNet has wider scope of mAP and Rank-1 performance on the three datasets. So it can be concluded that EnsembleNet has better generalization performance.

\section{Conclusion}
In this paper, we propose an ensemble network (EnsembleNet) and explored its priority. The designed EnsembleNet is based on ResNet-50. The features extracted from the multiple branches are concatenated as the final representation of a pedestrian. On three large-scale person re-ID datasets, experimental results show that our ensemble network achieves the state-of-the-art performance. And we also analyze the factors of the contributions. In the future, we try to combine the attention model into our work.
\section*{Acknowledgment}
This work has been supported by the National Natural Science Foundation of China (61806220).
%\begin{acks}
%   The work is
%  supported by the \grantsponsor{GS501100001809}{National Natural
%    Science Foundation of
%    China}{http://dx.doi.org/10.13039/501100001809} under Grant
%  No.:~\grantnum{GS501100001809}{61402519}.
%
%\end{acks}

\bibliographystyle{MyACMBib}
\bibliography{EnsembleNet2018}

\end{document}